\documentclass{article}
\PassOptionsToPackage{numbers, compress}{natbib}
\usepackage[preprint]{neurips_2022}
\usepackage[utf8]{inputenc} 
\usepackage[T1]{fontenc}    
\usepackage{hyperref}       
\usepackage{url}            
\usepackage{tabularx}
\usepackage{longtable}
\usepackage{booktabs}
\usepackage{booktabs}       
\usepackage{amsfonts}       
\usepackage{nicefrac}       
\usepackage{microtype}      
\usepackage{xcolor}         
\usepackage{multirow}
\usepackage{graphicx}
\usepackage{booktabs}
\usepackage{quoting}
\usepackage{float}
\usepackage{siunitx}
\usepackage{makecell} 
\usepackage{svg}
\usepackage{tcolorbox}
\usepackage{xcolor}
\usepackage{soul}

\tcbset{
  abstractstyle/.style={
    enhanced, 
    colback=green!5, 
    colframe=green!50!black, 
    fonttitle=\bfseries\large, 
    coltitle=black, 
    sharp corners=all, 
    boxrule=1pt, 
    arc=4mm, 
    left=6pt, 
    right=6pt, 
    top=6pt, 
    bottom=6pt, 
    width=\textwidth, 
    before skip=10pt, 
    after skip=10pt, 
  }
}

\usepackage{newtx}
\usepackage{caption} 
\usepackage[T1]{fontenc}

\usepackage{geometry}
\usepackage[]{mdframed}
\mdfsetup{%
middlelinecolor=red,
middlelinewidth=2pt,
backgroundcolor=gray!10,
roundcorner=10pt}

\usepackage[english]{babel}
\usepackage{hyperref}



\usepackage{appendix}
\usepackage{etoolbox}
\makeatletter
\patchcmd{\hyper@makecurrent}{%
    \ifx\Hy@param\Hy@chapterstring
        \let\Hy@param\Hy@chapapp
    \fi
}{%
    \iftoggle{inappendix}{
        \@checkappendixparam{chapter}%
        \@checkappendixparam{section}%
        \@checkappendixparam{subsection}%
        \@checkappendixparam{subsubsection}%
        \@checkappendixparam{paragraph}%
        \@checkappendixparam{subparagraph}%
    }{}%
}{}{\errmessage{failed to patch}}

\newcommand*{\@checkappendixparam}[1]{%
    \def\@checkappendixparamtmp{#1}%
    \ifx\Hy@param\@checkappendixparamtmp
        \let\Hy@param\Hy@appendixstring
    \fi
}
\makeatletter

\newtoggle{inappendix}
\togglefalse{inappendix}

\apptocmd{\appendix}{\toggletrue{inappendix}}{}{\errmessage{failed to patch}}
\apptocmd{\subappendices}{\toggletrue{inappendix}}{}{\errmessage{failed to patch}}

\hypersetup{colorlinks, linkcolor={red!55!black}, citecolor={blue!65!black}, urlcolor={blue!65!black}}

\title{Benchmarking the Pedagogical Knowledge of Large Language Models}

\author{
  Maxime Leli\`evre\textsuperscript{1,2} \And
  Amy Waldock\textsuperscript{1,2} \And
  Meng Liu\textsuperscript{1,2} \And
  Natalia Valdes Aspillaga\textsuperscript{1,2} \AND
  Alasdair Mackintosh\textsuperscript{1,2} \And
  Mar\'{i}a Jos\'{e} Ogando Portela\textsuperscript{1,2} \And
  Jared Lee\textsuperscript{1,2} \AND
  Paul Atherton\textsuperscript{1,2} \And
  Robin A. A. Ince\textsuperscript{1,2} \And
  Oliver G. B. Garrod\textsuperscript{1,2} \AND
  \normalfont\normalsize
  \textsuperscript{1}Fab Inc \\
  \textsuperscript{2}AI-for-Education.org
}

\begin{document}

\maketitle

\begin{abstract}

Benchmarks like Massive Multitask Language Understanding (MMLU) have played a pivotal role in evaluating AI's knowledge and abilities across diverse domains. However, existing benchmarks predominantly focus on content knowledge, leaving a critical gap in assessing models' understanding of pedagogy --- the method and practice of teaching. This paper introduces The Pedagogy Benchmark, a novel dataset designed to evaluate large language models on their Cross-Domain Pedagogical Knowledge (CDPK) and Special Education Needs and Disability (SEND) pedagogical knowledge. These benchmarks are built on a carefully curated set of questions sourced from the Chilean Ministry of Education's professional development exams for teachers, which cover a range of pedagogical subdomains such as teaching strategies and assessment methods. Here we outline the methodology and development of these benchmarks. We report results for 97 models, with accuracies spanning a range from 28\% to 89\% on the pedagogical knowledge questions. We consider the relationship between cost and accuracy and chart the progression of the Pareto value frontier over time. We provide online leaderboards at \url{https://rebrand.ly/pedagogy}, which are updated with new models and allow interactive exploration and filtering based on various model properties, such as cost per token and open-vs-closed weights, as well as looking at performance in different subjects. 

LLMs and generative AI have tremendous potential to influence education and help to address the global learning crisis. Education-focused benchmarks are crucial to measure models’ capacities to understand pedagogical concepts, respond appropriately to learners’ needs, and support effective teaching practices across diverse contexts. They are needed for informing the responsible and evidence-based deployment of LLMs and LLM-based tools in educational settings, and for guiding both development and policy decisions.

\end{abstract}

\newcommand{\dataurl}{https://huggingface.co/AI-for-Education/pedagogy-benchmark}
\newcommand{\codeurl}{https://github.com/AI-for-Education/pedagogy-benchmark}


\section{Introduction}

There is a global learning crisis, with 90\% of children in low-income countries unable to read a simple sentence by aged 10\footnote{\url{https://ieg.worldbankgroup.org/blog/back-basics-lessons-confronting-learning-crisis}}. The recent advances of large language models (LLMs) have brought new possibilities to education, with LLM-based educational tools being developed at a rapid pace \cite{wang2024large}, such as Khanmigo by Khan Academy and MagicSchool.ai. Although high-income countries have led the early adoption of these tools, there is also growing exploration of their application in low- and middle-income country contexts.  At AI-for-Education we have mapped\footnote{\url{https://ai-for-education.org/ai-products/}} over 280 AI-based EdTech products covering a range of application areas, such as adaptive personalized learning, student-facing chatbots, teacher assistants and learning management systems. These applications highlight the potential of LLMs to address educational challenges, including equitable access to quality learning, personalization of instructional approaches and reduction of teachers' workload. However, in this fast-moving landscape it is difficult to evaluate models from an education lens, as there are relatively few public benchmarks or evaluations which directly focus on educational applications. 

As generative models, LLMs can produce rich and varied outputs, which means that evaluating such models can be difficult. Much effort has been devoted to quantitatively assessing the performance of LLMs with \textit{benchmarks}: systematic quantitative evaluations of models' knowledge or abilities. These benchmarks typically consist of a dataset of questions or problems, together with some validated ground-truth or scoring function, so they can be applied to different models to obtain a numerical score for each. This score is intended to encapsulate some relevant aspect of model performance, and to facilitate direct comparison between models on that aspect. Many benchmarks have focused on evaluating the knowledge that LLMs have been able to encapsulate from their training corpus. For example, MMLU \cite{hendrycks2020measuring} sources multiple-choice questions from a wide range of human-focused exams. As models have increased in size and their knowledge base expands, performance on some existing benchmarks has saturated, leading to the development of more difficult tests, such as GPQA \cite{rein2023gpqa}, and more recently Humanity's Last Exam \cite{phan2025hle}. Both of these present extremely difficult technical questions that very few human experts could answer. Such benchmarks have been widely employed to compare models, and model developers typically publish scores from popular benchmarks when they release new models. Multiple leaderboards and comparison sites provide systematic comparisons of available models\footnote{For example, \url{https://livebench.ai}, \url{https://huggingface.co/open-llm-leaderboard}, \url{https://www.vellum.ai/llm-leaderboard} to list just a few.}. 

From an education perspective, it is important to distinguish between content knowledge (the factual or conceptual understanding of a subject) and pedagogical knowledge (understanding the methods and strategies used to effectively teach a subject). After reviewing the LLM benchmark literature we did not find any existing LLM benchmarks which explicitly focus on knowledge of pedagogy, such as understanding of learning theories, instructional design and teaching practices. With the rapid rise in educational applications, we sought to address this gap by developing a novel Pedagogy Benchmark. This is a set of 920 multiple choice questions (MCQs) from teacher training exams, which evaluate LLM's cross-domain pedagogical knowledge (CDPK). We also provide a benchmark focusing on Special Educational Needs and Disability (SEND), which consists of a further 223 questions focussed on this area. 

By bridging this gap, our benchmarks aim to accelerate and improve the responsible development of LLMs for educational applications, paving the way for more impactful and evidence-based applications of AI in education. Here we describe the development of these benchmarks together with a comprehensive set of results for 97 current and historical LLMs covering a range of providers and model sizes. We have made the data\footnote{\url{\dataurl}} and code for running the benchmarks openly available\footnote{\url{\codeurl}} and maintain regularly updated leaderboards showcasing the results\footnote{\url{https://rebrand.ly/pedagogy}}. 


\section{Related Work}
\label{sec:related_work}

Zhao et al. \cite{zhao_survey_2023} provide a taxonomy of LLM abilities, considering at the highest level: language generation, knowledge utilisation, complex reasoning, and other advanced abilities. Language generation can include open-ended tasks like text summarization or code generation, while advanced abilities can include areas like human-alignment and tool use. Different abilities in this taxonomy present different challenges for the development of quantitative benchmark assessments. 

For example, language generation tasks are fundamentally open-ended and subjective, often without clear correct or incorrect answers. While some code outputs can be run and objectively tested for correct functionality, other open-ended language tasks are more difficult to assess. Benchmarks in this area can use various approaches to score outputs generated from a standardised dataset of prompts, such as heuristic answer matching, human ratings (which can be difficult or costly to obtain at scale) or an evaluation by other LLMs (often termed \emph{LLM-as-a-judge} \cite{zheng2023judging}). An example of a language generation benchmark that uses human ratings at scale is Google's recent Gemini evaluations \cite{team2025evaluating} (discussed below) which uses 395 human experts to generate and review thousands of human-LLM chatbot interactions. An example of using LLM-as-a-judge at scale is HealthBench \cite{arora_healthbench_2025} from OpenAI, which considers model responses to 5000 different multi-turn conversations related to healthcare. 262 human experts (qualified physicians) generated more than 48,000 example-specific binary evaluation rubrics, which are then evaluated using LLM judges. Another example of the LLM-as-a-judge approach being applied in the education domain is the work of Oak National Academy, which uses 24 different rubrics to assess and compare AI-generated lesson plans \cite{clark_auto-evaluation_2025}. 

An alternative approach is to use a form of continuous user-blind A-B testing, in which real users provide queries on any topic, are presented with answers from two or more unlabelled models and are asked to choose their favourite. Using algorithms developed for ranking players in sports like chess, a score is computed for each model based on its aggregate performance across a range of head-to-head comparisons. An example of this is the Chatbot Arena \cite{chiang2024chatbot}, which provides a leaderboard based on user preference in response to user-provided queries. This approach is interesting, but presents some challenges including obtaining enough users with diverse queries, the ongoing costs of hosting and inference, and possible systematic changes in users' queries and preferences over time. One particular criticism is that scoring in this system can be sensitive to relatively superficial stylistic properties of the output, like the use of structured headings, bullet list, emojis etc. which have led to attempts to control such rankings for length and style\footnote{\url{https://lmsys.org/blog/2024-08-28-style-control/}}.

\subsection{Existing LLM Knowledge Benchmarks}

We focus primarily on knowledge utilization, as our specific aim is to test models' pedagogical knowledge. For benchmarks in the area of knowledge utilization, the dataset of questions is core to the evaluation process and can consist of either closed- or open-ended questions. Closed-ended question formats, such as multiple-choice questions, make the output of the LLM easier to parse and mark as they feature an answer that is easy to extract, match to the correct answer, and aggregate over the full set of questions to give an accuracy score. For this reason, the MCQ approach has been the most common approach used for LLM knowledge benchmarks. 

The Massive Multitask Language Understanding (MMLU) benchmark \cite{hendrycks2020measuring}, has received much attention as a tool to compare models' knowledge. MMLU consists of 15,908 questions across 57 subjects, drawn from human exams across a range of topics, such as STEM fields, international law and nutrition, and across different levels from primary and secondary school through to professional examinations (e.g. Examination for Professional Practice in Psychology). However, as models have improved over time performance on MMLU has started to saturate, and there is a risk that due to it's popularity the open-data questions would appear in training data or be used as a target for explicit fine-tuning. To address this MMLU-Pro \cite{wang2024mmlu} has been developed as a more robust and difficult alternative. This selects harder questions, with a larger number of response options and includes more questions that require reasoning rather than just knowledge retrieval. In general, the rapid pace of model development has meant this issue of saturation has been a common one, and it is an increasing challenge to develop benchmarks that are hard enough to challenge current models. 

The Google-Proof Q\&A (GQPA) benchmark \cite{rein2023gpqa}, was another attempt to address this problem. This is a set of 448 specially written multiple-choice questions which were designed to be extremely difficult, requiring in-depth expert (e.g. PhD-level) knowledge that is not easily available from web searches. However, despite the difficulty of the questions, performance of new models has rapidly increased in the few years since it was released. Currently, the best model scores more than 85\% accuracy on GPQA-Diamond, a subset of the most challenging and carefully verified questions. Humanity's Last Exam \cite{phan2025hle} attempts to develop questions with even higher levels of difficulty, and addresses this through a crowd-sourcing procedure where experts are invited to submit their most challenging questions. Humanity's Last Exam consists of 2500 questions, and is by some margin the hardest such multiple-choice benchmark, with current best accuracy around 21\%. However, the high-level of the difficulty of the questions goes beyond assessing content knowledge, as many questions require scientific and logical reasoning to derive the answer. 

Despite this tendency for models to saturate knowledge benchmarks relatively quickly after they are developed, we nonetheless consider it important to develop a benchmark of pedagogical knowledge to fill the gap we identified in the current landscape. Even if performance on our new benchmark saturates for newer models, it is still useful to track performance for smaller, open-source, offline and on-device models, particularly in the educational context of low- and middle-income countries. 

\subsection{Existing Education-focused LLM Benchmarks}

In a recent note, AI-for-Education mapped existing benchmarks focusing on four specific educational use-cases\footnote{\url{https://rebrand.ly/9vl3pha}}: lesson plan generation, adaptive student exercises, student question answering, and assessment generation. There were relatively few directly relevant benchmarks with publicly available data that could be run automatically on new models (i.e. with code available and without requiring human input or ratings). These included benchmarks related to pedagogical alignment of student interactions \cite{macina2025mathtutorbench}, identifying specific misconceptions underlying student errors in maths problems \cite{nancy2024mathmisconceptions} and generating multiple-choice questions \cite{bitew2023distractorgeneration}. Khan Academy has released the Conversation-Based Maths Tutoring Accuracy (CoMTA) Dataset, which consists of a data set of student-chatbot interactions focused on maths problems, in which the challenge is for the model to determine whether the student's attempted answer in the final turn is accurate or not \cite{miller_llm_2024}. MinorBench \cite{khoo2025minorbench} evaluates LLMs on their ability to refuse unsafe or inappropriate queries from children.

There have also been a series of papers from Google's LearnLM Team exploring approaches to evaluating LLMs for educational applications in depth. Jurenka et al. \cite{jurenka2024towards} consider the problem of assessing the pedagogical practice of an LLM tutor. Pedagogical practice refers to the actual behaviour of the tutor, and whether that behaviour embodies sound pedagogical principles, for example, not revealing the answer to a problem too early. However, this is challenging and requires manual ratings of real multi-turn student-chatbot interactions by human experts. In \cite{team2024learnlm}, the LearnLM team focus on pedagogical instruction following, with assessment again based on a large pool of pedagogy experts analysing and rating real conversation traces. To give an idea of the scale of this undertaking, they obtain 10,192 assessments of 2360 conversations consisting of over 58,000 messages from 228 experts. These interactions are rated against pedagogical principles such as managing cognitive load, inspiring active learning, deepening metacognition, stimulating curiosity and adapting to the learner. Most recently \cite{team2025evaluating}, they adopt an arena style head-to-head in which pedagogical experts role playing specific student personas interact with a pair of models side-by-side, before selecting their preference. These interactions are then further rated by a different set of experts. While this is extremely promising and thorough approach, the cost and difficulty involved with obtaining expert ratings of large-scale real human-LLM interactions, mean it is not possible to continually test a wide range of models in this way. 

In our literature review, we have not found any existing benchmarks that explicitly test pedagogical knowledge in an easily replicable way that can be directly applied to new models.


\section{Methods}

\subsection{Source of Questions}

We extracted multiple-choice questions (MCQs) from the Chilean Ministry of Education's ECEP program. The ECEP (\emph{Evaluación de Conocimientos Específicos y Pedagógicos}, Evaluation of Specific and Pedagogical Knowledge), is an annual standardised test in Chile designed to assess teachers’ mastery of both content and pedagogical knowledge across various education levels and subjects.  

Administered by the Agency for Education Quality\footnote{\url{https://www.agenciaeducacion.cl/}} (\emph{Agencia de Calidad de la Educación}), and developed by the Education Quality Agency and Center for Pedagogical Improvement, Experimentation, and Research of the Chilean Ministry of Education\footnote{\url{https://www.cpeip.cl/cpeip/}} (\emph{Agencia de la Calidad de Educación y Centro de Perfeccionamiento, Experimentación e Investigaciones Pedagógicas}) the assessment instrument comprises 60 multiple-choice questions distributed across domains -- including pedagogical knowledge --  and with varying levels of difficulty and complexity. The results of the ECEP, together with the Teaching Portfolio, are used to determine teachers’ progression within the Teacher Professional Development and Recognition System (\emph{Sistema de Reconocimiento del Desarrollo Profesional Docente}). This system supports, recognises and promotes career advancement by guiding teachers towards an expected level of professional development and providing an incentive pathway to continue their growth and effectiveness in the classroom\footnote{Article 19, Law No. 20.903 (2016), Establishing the Teaching Career System, Chile. }.

The Teacher Professional Development and Recognition System applies to teachers working in municipal, government-subsidised private schools, and other publicly funded educational institutions. The assessments are given substantial weight particularly for early career teachers, with failure to pass after 2 attempts potentially resulting in dismissal or the loss of seniority within the system.\footnote{Congreso Nacional de Chile. (2023). Ley N° 21.625: Establece sistema único de evaluación docente. Diario Oficial de la República de Chile, 24 de octubre de 2023. Accesed at: \url{https://www.bcn.cl/leychile/navegar?idNorma=1197193}. See also \url{https://cpeip.cl/sistema-reconocimiento/}}.

We sourced questions from ECEP exam papers administered between 2017–2023\footnote{\url{https://www.cpeip.cl/ya-estan-disponibles-las-pruebas-ecep-2017-al-2020/}}across all 7 educational levels: early childhood (\emph{Educación Parvularia}, EP), early primary (\emph{Primer Ciclo Generalista}, PC),  later primary (\emph{Educación Básicas Segundo Ciclo}, SC), secondary (\emph{Educación Media}, EM), special education (\emph{Educación Especial } EE; \emph{Educación Diferencial}, ED), adult education (\emph{Educación de Adultos}, EA), and vocational/technical secondary education (\emph{Educación Media Técnico Profesional}, EMTP). Note that for the higher levels there are specific exams for different subject areas, such as mathematics, history, biology, etc.  

\begin{figure}[htbp]
  \centering
  \includesvg[width=\textwidth]{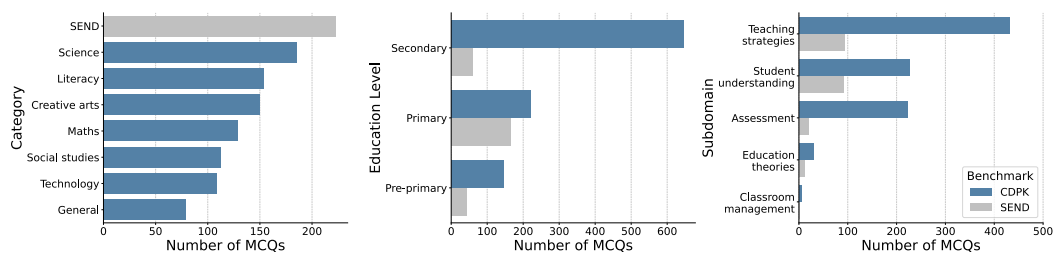}
  \caption{Distribution of the MCQs by Category (left), Education Level (center) and Education Subdomain (right). Note that questions could be tagged with more than one education level.}
  \label{fig:dataset_overview}
\end{figure}

Our pipeline involved identifying, extracting, cleaning and translating questions from the PDF files for each exam. Correct answers were provided in separate documents, which were parsed and matched to the extracted questions. The initial extraction resulted in a pool of 9,978 MCQs obtained from 192 different exams (across different levels, subjects and years). This pool of MCQs was de-duplicated and manually curated by a pedagogical expert to ensure quality, relevance and correctness. Each question had the associated metadata in terms of level, from pre-primary to secondary, and subject area obtained from the topic of the exam containing the question. These were manually checked by the pedagogical expert and were amended as necessary (see Fig. \ref{fig:dataset_overview}). In addition, the pedagogical expert labelled each question with a pedagogical subdomain (e.g. teaching strategies, student understanding, assessment, education theories, classroom management). This resulted in 1143 MCQs labelled with education level, subject and pedagogical domain. These questions are available at \url{\dataurl}.

We considered the 223 questions from the category Special Education Needs and Disability (SEND) separately from the others (Section \ref{sec:send}). SEND questions were separated because they involve specialist knowledge as well as teaching and assessment strategies not typically expected for mainstream classroom teachers. While some SEND needs are addressed in regular schools, many of the scenarios in these questions refer to cases managed by specialist staff or in special schools. This distinction is also reflected in teacher training, where SEND is often treated as a separate course or stream. Consequently, we considered it valuable to test these questions separately and report a distinct SEND-specific knowledge score. Note that all questions used in the SEND benchmark came from the exams covering special education ((\emph{Educación Especial} EE; \emph{Educación Diferencial}, ED). The remaining 920 questions formed our primary cross-domain pedagogical knowledge benchmark.

\begin{table}[htbp]
    \centering
    \fontsize{8}{9}\selectfont
    \begin{tabularx}{\textwidth}{X}
    
        \toprule
        \multicolumn{1}{c}{\textbf{Science (Pre-primary)}} \\
        \midrule
        At a preschool level, the educational team needs to evaluate in everyday contexts the following learning objective from the Natural Environment Exploration core: "Communicate some basic properties of the natural elements they explore, such as colors, textures, sizes, temperatures among others". Which performance of the children would demonstrate the achievement of the selected learning objective? \\
        \\
        A) During a learning experience, Santiago draws stones of different sizes\\ 
        B) During lunchtime, Pablo looks at the spoon and seeing his reflection says: "Look, Pablo!"\\ 
        C) During an experience in the yard, Maria collects different types of tree leaves in a basket\\ 
        D) During free play time in the yard, Soledad rubs some seeds in her hands and says: "Prickly"\\ 
        \\
        Correct answer: D \\

        \toprule
        \multicolumn{1}{c}{\textbf{Social studies (Primary)}} \\
        \midrule
        According to the following objective: "Read simple plans of your environment, using reference points and categories of relative position," what activities are most appropriate to evaluate it in Year 2? \\
        \\
        A) Draw a plan of the classroom \\ 
        B) Interpret the symbols of a city plan \\ 
        C) Describe the location of places on a given map \\ 
        D) Determine the coordinates of given points on a grid \\ 
        \\
        Correct answer: C \\

        \toprule
        \multicolumn{1}{c}{\textbf{General (Secondary)}} \\
        \midrule
        According to Bowlby's theory, which of the following statements corresponds to a characteristic of attachment? \\
        \\
        A) It is a process that lasts until the early years of life \\
        B) It is a bond that must be stable and continuous to ensure proper development \\
        C) It is a bond that determines all affective relationships in life and development \\
        D) It is the first relationship of the newborn with its mother and cannot be replaced by another person \\
        \\
        Correct answer: B \\
        \bottomrule
    \end{tabularx}
    \caption{Example questions from the dataset with different subject categories and education level.}
    \label{tab:example_mcqs}
\end{table}

\subsection{Question Processing}

\begin{figure}[htbp]
  \centering
  \includegraphics[width=\textwidth]{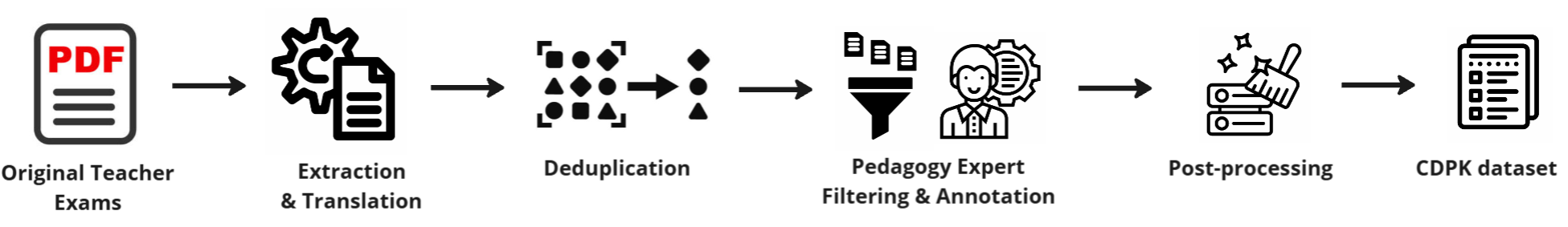}
  \caption{Dataset Construction Pipeline}
  \label{fig:flowchart}
\end{figure}

Figure \ref{fig:flowchart} shows a schematic of our question processing pipeline, which is detailed below. 

\subsubsection{Question Extraction and Pre-processing}

\textbf{Initial Extraction:} The first step involved extracting the questions from the exam PDFs. Given that the layout of the PDFs varied across different years and subjects, some care was necessary to manually adjust the page range for extraction of each exam, and regular expressions were used to identify the questions.  

The PDF files for the exam years 2021, 2022, and 2023 did not contain embedded text which could be extracted programmatically (as they were scanned documents rather than vector text). For these files, an Optical Character Recognition (OCR) step was incorporated using GPT-4o-mini (see prompt used in \ref{prompt_ocr}).
    
\textbf{Data Structuring and Translation:} GPT-4o-mini was used to process and translate the extracted Spanish language question texts. Processing each question separately, the model was prompted to translate to English and isolate the stem and answer options (see prompt used in \ref{prompt_data_extraction}). The structured outputs function of OpenAI’s API ensured that all MCQs contained a well-defined stem and four options. For each PDF, the translation of a small random sample of MCQs were manually checked by a human proficient in English and Spanish.

\textbf{Answer Extraction:} The correct answers were manually extracted from separate answer key PDFs documents. 

\subsubsection{De-duplication}

\begin{figure}[tph]
  \centering
  \includesvg[width=0.5\textwidth]{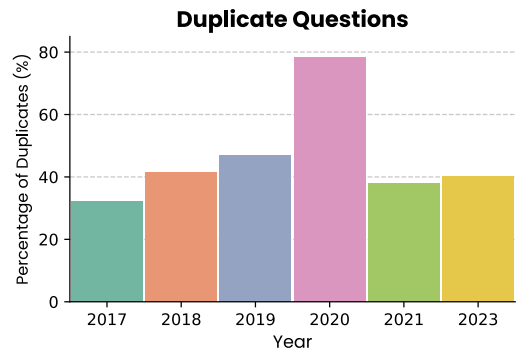}
  \caption{Percentage of duplicate questions (questions in each year that appeared in another year). Note that 2020 was an outlier year where the majority of questions were repeated.}
  \label{fig:duplicates_distrib}
\end{figure}

To prevent the inclusion of duplicate questions across multiple years, we implemented a de-duplication pipeline using fuzzy matching based on Levenshtein Distance. The Ministry of Education had noted on its website that questions would be reused across years, but provided no specific numbers or further details. Here, we labelled a sample as a duplicate if either the stem alone, or the whole question (stem and answer options) were above the fuzzy similarity ratio of 90 (where a score of 100 corresponds to a perfect match). Both potential within-year (between exam) and cross-year duplicates were addressed during this process. This de-duplication significantly reduced the number of MCQs, as it was found that many years contained more than 30\% duplicate questions (see Fig. \ref{fig:duplicates_distrib}). 

\subsubsection{Human Review and Annotation}

To ensure the quality and relevance of the MCQs, each question was manual inspected and annotated by a pedagogy expert. The expert reviewed each MCQ and edited them based on the following criteria: 
\begin{itemize}
    \item Relevance to pedagogy: the question should assess some pedagogical knowledge. Questions that were too focused on subject-specific content knowledge without educational context were excluded.
    \item Clarity and correctness: ambiguous, poorly formatted, incomplete or badly translated MCQs were excluded.
    \item Context agnostic: MCQs specific to the Chilean context or requiring specific knowledge about Chile (for example about specific legislation, or local culture) were excluded.
\end{itemize}

Where appropriate and necessary, the pedagogy expert made minor edits to MCQs with changes including translating some terms to reflect international norms and correcting punctuation or grammatical errors in the translation.

Once this first quality check passed, the pedagogy expert tagged each selected MCQ with the subject category (e.g. Literacy, Math, History, SEND, ...), the education level (e.g. Primary, Secondary,...) and the pedagogical domain (e.g. teaching strategies, classroom management, assessment, ...) (see Figure \ref{fig:dataset_overview}).

\subsubsection{Post-Processing and Final Filtering}

After running the initial dataset through 36 different LLMs, it was observed that approximately 10\% of the MCQs were consistently answered incorrectly by most models (31 out of 36 LLMs), with some never being answered correctly at all. This suggested potential issues with the quality of those questions. Upon closer manual inspection, several flaws were identified in these MCQs, including grammatical errors in the stem or options, multiple possible correct answers or ambiguity in the question or options. These MCQs were excluded from the final dataset.

Following these steps, the final dataset included 1143 high-quality, annotated MCQs, which have been made publicly available for benchmarking the pedagogical knowledge of LLMs\footnote{\url{\dataurl}}.

\subsection{Benchmark Evaluation}

Our evaluation pipeline consists of a carefully developed prompt and response parsing script as well as a complete open-source framework allowing to run a wide range of models from different providers such as OpenAI, Anthropic, OpenRouter, Fireworks.ai, Amazon Bedrock and others. This is based on Fabdata-LLM\footnote{\url{https://github.com/AI-for-Education/fabdata-llm}} developed under the AI-for-Education initiative\footnote{\url{https://ai-for-education.org/}}, which provides a common Python interface to a range of different LLM providers. Code implementing the benchmark procedure is available at \url{\codeurl}.

Following common methodologies and best practices in the LLM benchmarks field \cite{brown2020language, wang2024my, mcintosh2024inadequacies, biderman2024lessons, laskar2024systematic, chang2024survey, guo2023evaluating}, we used a few-shot prompting technique, and used the same prompt for all models. The prompt template (full example can be seen in \ref{prompt_example}) used for the evaluation was:

{\centering
\begin{tcolorbox}[title=Prompt Template,width=0.85\textwidth]
    \textbf{The following are example multiple choice questions (with answers).}\\
    Question and options 1 \\ 
    Correct answer 1 \\
    Question and options  2 \\
    Correct answer 2 \\
    Question and options 3 \\
    Correct answer 3 \\
    \\
    \textbf{Answer the following real question using same answer format:} \\
    Question and options (to answer) \\
    \\
    \textbf{Only answer the real question.} \\
    \textbf{Only provide the letter for your answer.} \\
    \textbf{Stop exactly after the letter.}
    \label{prompt_template}
\end{tcolorbox}
}

Three fixed few-shot examples were selected at random within each subject area. All questions from that subject area used the same few-shot examples, and those three questions were not used in the benchmark. There were seven subject areas, and so out of the total of 920 general pedagogy MCQs, 21 were used as few-shot examples and the CDPK benchmark accuracy is determined over 899 tested questions. For SEND there were 223 MCQs, and the benchmark accuracy is determined over 220 tested questions. 

We conducted preliminary experiments to test various configurations of the prompt and response parsing algorithms. First, this iterative process involved adjustments to the phrasing of instructions to improve clarity and reduce ambiguity. Initial experiments with first versions of the prompt showed that it was working well for state-of-the-art models which were able to follow instructions precisely, but less well for smaller models with reduced instruction following capabilities. Second, it involved adjustments to the parsing logic to handle edge cases such as incomplete or badly formatted outputs. Again, early experiments with the parsing script showed a negative bias for models with few parameters, preventing them from being evaluated correctly. The parsing script was thus implemented to handle deviations from the strict expected format mentioned in the prompt so that outputs such as "Answer A", "Answer: A" or "Correct Answer: A", where a clear answer choice could be unambiguously extracted, would be included. Through these experiments, we identified a prompt and a parsing script that worked consistently across a diverse set of models, especially of different sizes and capabilities.

Model performance was assessed using overall accuracy (percentage of correctly answered questions). Specifically, we computed the exact match rate between the model’s selected answer letter (after being parsed as described above) and the correct answer. Models exhibiting more than 5\% of badly formatted responses (responses where a clear choice option could not be extracted) were considered to be unable to perform the benchmark. The results of these models were not considered further, and they are excluded from the leaderboard. 

LLMs often employ stochastic decoding strategies (e.g., token sampling or beam search), which can lead to variability in their outputs. To evaluate the stability of our benchmark given stochastic LLM responses we performed an experiment in which we ran the 899 CDPK questions 20 times. To limit the computational cost this was done with a subset of 8 models sampled across the range of obtained accuracies. Over these 8 models the maximum standard deviation of accuracy (over the 20 runs) was 0.43\%. This shows that our test procedure and result scores are robust and repeatable. 

In the results below we report accuracies with 95\% bootstrap confidence intervals obtained from resampling questions with replacement 1000 times, separately for each model. 

\subsection{Answer position selection bias}

It has been shown that LLMs can exhibit a \textit{selection bias} when evaluated on multiple-choice questions \cite{zheng2023large,wang2023large,pezeshkpour2023large,gupta2024changing}. The main causes being reported are a \textit{token-level bias} and  \textit{positional bias}. Token-level bias refers to when models favour particular answer tokens due to their overall probability of occurrence, for example the token "A" might appear more frequently in English than the token "D". Positional bias refers to a potential bias to favour the answer in a particular position presented options, i.e. favouring the answer listed last as it is most recent in the context window. 

To test for these effects, we ran a second experiment with the CDPK dataset. We generated four configurations where the correct answer appeared in each possible position (A, B, C, D). In each case, the correct answer was swapped with the original answer in that place so ordering of the two other options was preserved. 

\begin{figure}[htpb]
  \centering
  \includesvg[width=\textwidth]{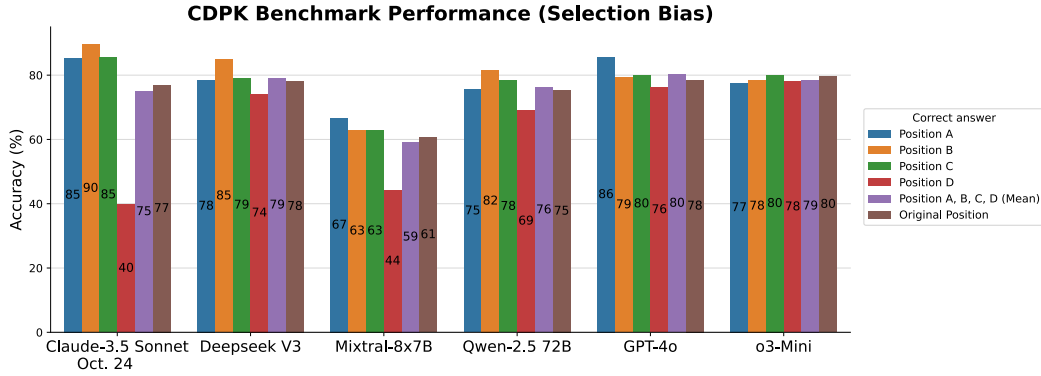}
  \caption{Test of a subset of models for a selection bias. In particular, some models exhibit a selection bias when the correct answer is the index D (and in 4th position).}
  \label{fig:plot_selection_bias}
\end{figure}

We indeed found a selection bias for some of the models tested (see Fig. \ref{fig:plot_selection_bias}) but contrary to \cite{zheng2023large} who showed a preference for response \emph{A}, we found that some models exhibit a negative bias against selecting option \emph{D}. However, this approach cannot separate token-level and positional bias. In the CDPK dataset, the correct answers are approximately balanced, and our main single-run results (brown) are very close to the average of the 4 different answer position runs (purple). 

\section{Results}

\subsection{Cross-Domain Pedagogical Knowledge (CDPK) Benchmark}

\begin{figure}[htpb]
  \centering
  \includesvg[width=\textwidth]{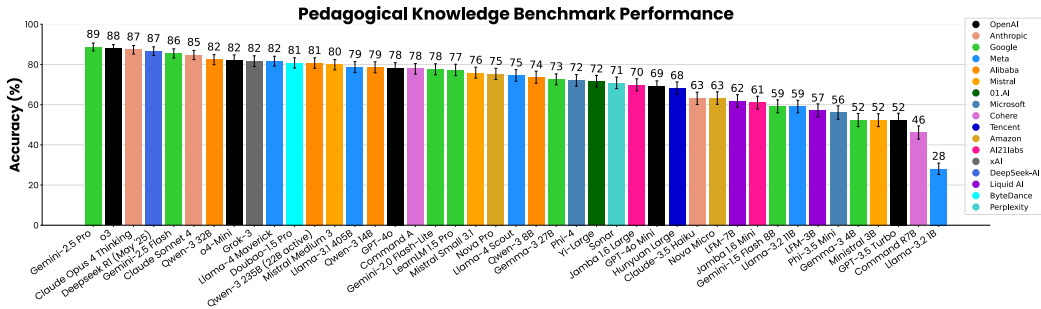}
  \caption{Accuracy on the Pedagogy Benchmark (CDPK) for a subset of models. Error bars show 95\% bootstrap confidence intervals}
  \label{fig:cdpk_results_overall}
\end{figure}

At the time of the writing, 97 LLMs have been tested on The Pedagogy Benchmark, including closed- and open-source models,  from a range of providers, and covering a wide range of model sizes and inference costs. Figure \ref{fig:cdpk_results_overall} and Table \ref{tab:cdpk_accuracy_subset} show the performance of a subset of models. All results can be found in Appendix \ref{sec:full_results} (Table \ref{cdpk_accuracy_full}).

\begin{figure}[htpb]
  \centering
  \includesvg[width=\textwidth]{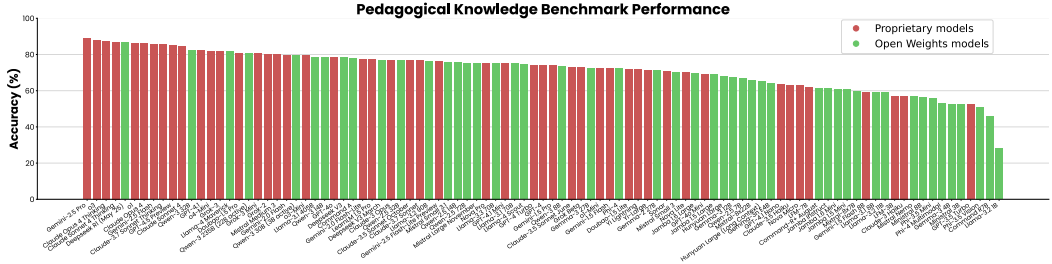}
  \caption{Accuracy on the Pedagogy Benchmark (CDPK) by Weights Availability.}
  \label{fig:cdpk_results_open}
\end{figure}

\begin{figure}[htpb]
  \centering
  \includesvg[width=\textwidth]{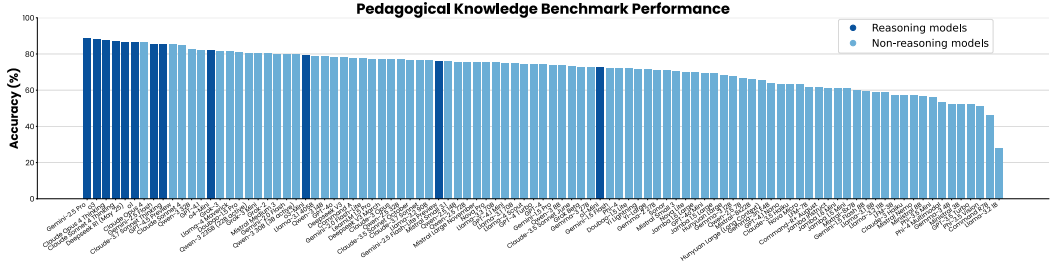}
  \caption{Accuracy on the Pedagogy Benchmark (CDPK) by Reasoning Capability}
  \label{fig:cdpk_results_reasoning}
\end{figure}

Accuracies range from 28\% (Llama-3.2 1B) to 89\% for the current leader, Google's Gemini 2.5 Pro model. Closed-source models dominate at the top of the leaderboard (Figure \ref{fig:cdpk_results_open}), with 9 out of the top 10 being closed models from OpenAI, Google and Anthropic. However, the fifth-place model with 87\% accuracy is the open-source Deepseek R1 (May 2025 release). It is also noteworthy that the many of the top-models are \emph{thinking} models (Figure \ref{fig:cdpk_results_reasoning}) which are fine-tuned for inference-time chain of thought reasoning.

\begin{table}[htpb]
    \centering
    \resizebox{\textwidth}{!}{%
    \begin{tabular}{l c l| l c l}
    \toprule
    Model & Accuracy & Company & Model & Accuracy & Company \\
    \midrule
    Gemini-2.5 Pro & 88.77 & \textcolor{gray}{Google} & Qwen-3 8B & 73.53 & \textcolor{gray}{Alibaba} \\
    o3 & 87.88 & \textcolor{gray}{OpenAI} & Gemma-3 27B & 72.64 & \textcolor{gray}{Google} \\
    Claude Opus 4 Thinking (low) & 87.43 & \textcolor{gray}{Anthropic} & Phi-4 & 72.19 & \textcolor{gray}{Microsoft} \\
    Deepseek R1 (May '25) & 86.65 & \textcolor{gray}{DeepSeek-AI} & Yi-Large & 71.52 & \textcolor{gray}{01.AI} \\
    Gemini-2.5 Flash & 85.54 & \textcolor{gray}{Google} & Sonar & 70.75 & \textcolor{gray}{Perplexity} \\
    Claude Sonnet 4 & 84.76 & \textcolor{gray}{Anthropic} & Jamba 1.6 Large & 69.86 & \textcolor{gray}{AI21labs} \\
    Qwen-3 32B & 82.42 & \textcolor{gray}{Alibaba} & GPT-4o Mini & 69.19 & \textcolor{gray}{OpenAI} \\
    o4-Mini & 81.98 & \textcolor{gray}{OpenAI} & Hunyuan Large & 68.19 & \textcolor{gray}{Tencent} \\
    Grok-3 & 81.76 & \textcolor{gray}{xAI} & Claude-3.5 Haiku & 63.29 & \textcolor{gray}{Anthropic} \\
    Llama-4 Maverick & 81.65 & \textcolor{gray}{Meta} & Nova Micro & 63.18 & \textcolor{gray}{Amazon} \\
    Doubao-1.5 Pro & 80.76 & \textcolor{gray}{ByteDance} & LFM-7B & 61.85 & \textcolor{gray}{Liquid AI} \\
    Qwen-3 235B (22B active) & 80.65 & \textcolor{gray}{Alibaba} & Jamba 1.6 Mini & 61.07 & \textcolor{gray}{AI21labs} \\
    Mistral Medium 3 & 79.98 & \textcolor{gray}{Mistral} & Gemini-1.5 Flash 8B & 59.18 & \textcolor{gray}{Google} \\
    Llama-3.1 405B & 78.75 & \textcolor{gray}{Meta} & Llama-3.2 11B & 59.07 & \textcolor{gray}{Meta} \\
    Qwen-3 14B & 78.53 & \textcolor{gray}{Alibaba} & LFM-3B & 57.06 & \textcolor{gray}{Liquid AI} \\
    GPT-4o & 78.31 & \textcolor{gray}{OpenAI} & Phi-3.5 Mini & 56.06 & \textcolor{gray}{Microsoft} \\
    Command A & 77.86 & \textcolor{gray}{Cohere} & Ministral 3B & 52.39 & \textcolor{gray}{Mistral} \\
    Gemini-2.0 Flash-Lite & 77.64 & \textcolor{gray}{Google} & Gemma-3 4B & 52.39 & \textcolor{gray}{Google} \\
    LearnLM 1.5 Pro & 77.31 & \textcolor{gray}{Google} & GPT-3.5 Turbo & 52.28 & \textcolor{gray}{OpenAI} \\
    Mistral Small 3.1 24B & 75.86 & \textcolor{gray}{Mistral} & Command R7B & 46.05 & \textcolor{gray}{Cohere} \\
    Nova Pro & 75.31 & \textcolor{gray}{Amazon} & Llama-3.2 1B & 28.03 & \textcolor{gray}{Meta} \\
    Llama-4 Scout & 74.53 & \textcolor{gray}{Meta} &  &  &  \\
    \bottomrule
    \end{tabular}
    }
    \caption{Accuracy scores on The Pedagogy Benchmark (for a subset of models), sorted from highest (top left) to lowest (bottom right).}
    \label{tab:cdpk_accuracy_subset}
\end{table}

\subsubsection{Estimated human baseline}

Based on the results of more than 25,000 teachers from 2017 to 2021, the mean expected score of a human trainee teacher would be expected to be approximately 50\% on The Pedagogy Benchmark. However, this is only an estimate, and any interpretation should be made with caution. Question-level human results are not available for these exams, so we cannot directly quantify human accuracy on the specific questions included in these benchmarks. Instead, we have estimated this based on aggregate subject- and year-level results by allocating each benchmark question the average accuracy of the paper from which it was drawn. We tested four different methods, all of which produced an estimated accuracy for human teachers on CDPK of around 50\%. It is also important to emphasize that teachers typically take exams only in the subjects they teach, not in all subjects. Therefore, evaluating overall human performance on the CDPK benchmark is difficult, as no individual teacher answers all its constituent questions.

\subsubsection{Accuracy vs Cost: The Value Frontier}

For each benchmarked model we also record the inference costs, where available. Note that occasionally experimental models are provided free of charge without any price information; such models do not appear in this value analysis. Inference costs are usually provided as USD per million tokens, and often have separate rates for input tokens (user prompt and context) and output tokens (model generated responses). Here we use just input token costs, as the intention was to benchmark knowledge and the prompt is set up to encourage models to answer directly. However, in the era of reasoning models, thinking traces are charged as output tokens, even if they are not directly accessible as the end user. For open-source models there are often multiple hosting providers available. This complicates benchmarking as different providers can be running different versions of the model (e.g. using different quantisations of the weights), or using other inference-time optimisations or shortcuts. We try to find the most reliable provider for any given model in its original quantisation, and report the input costs for that provider. This means sometimes models may be hosted elsewhere at lower cost than we report, but we cannot be sure the benchmark performance would be the same. 

\begin{figure}[htpb]
  \centering
  \includegraphics[width=\linewidth]{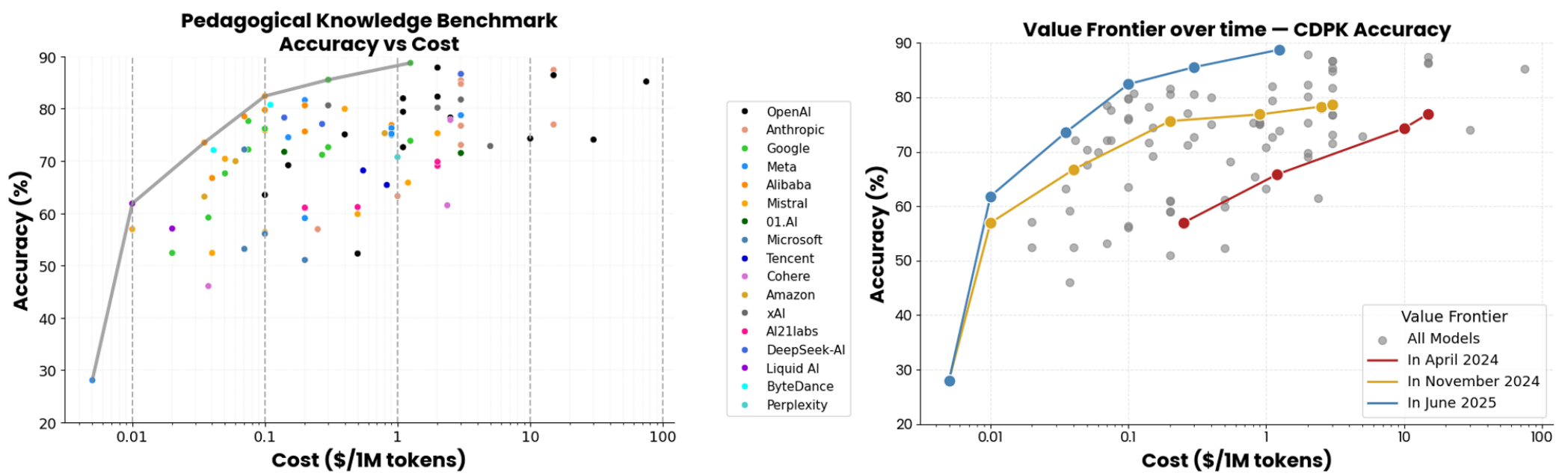}
  \caption{Left: Trade-off between performance and cost, the best models being on the \textit{value frontier} (grey line). Right:Value frontier over time.}
  \label{fig:scatters_cdpk_cost}
\end{figure}

Figure \ref{fig:scatters_cdpk_cost} shows these results. The left panel shows the current Pareto value frontier -- technically the convex upper hull. We can see there are models available at a wide range of inference costs, spanning nearly four orders of magnitude from 1¢ to nearly \$100 per million input tokens. It is also noteworthy that the most expensive model does not have the highest accuracy although the cheapest model is the worst performing. For any particular inference cost there is a range of around 20 percentage point difference in accuracy between the lowest and highest scores for models near that price point. 

The current value frontier is illustrated in grey and includes the following models (in increasing order of performance and cost): Llama 3.2 1B (Meta), LFM 7b (Liquid AI), Qwen 3 8B (Alibaba), Qwen 3 32B (Alibaba), Gemini 2.5 Flash (Google) and Gemini 2.5 Pro (Google). It's interesting that while the majority of the top performing models were proprietary, on the value-frontier more than half are open-weight (Llama and Qwen families). However, at higher costs, closed models are ahead. Google's latest family of models provide the best value above 10¢/Mtoken. Between 5-10¢/Mtoken Qwen 3 models provide best value, although it worth noting that models from ByteDance (Doubau 1.5, Lite and Pro) and Google (the previous generation of Gemini 2.0 Flash and Flash Lite) are very close to the frontier in this range. At the low end of cost it's interesting that a novel non-transformer-based architecture from Liquid AI (LFM-7B) performs very well (62\% accuracy for 1¢/Mtoken), although Mistral Nemo is not far below (57\% accuracy for the same price). 

Figure \ref{fig:scatters_cdpk_cost} (right panel) shows how this value frontier has evolved over time, showing the rapid pace of improvement and reduction in cost for all performance levels. For example, at \$0.10 per million tokens, the performance has increased from ~50\% (April 2024) to ~70\% (November 2024) to 82\% (June 2025). 82\% exceeds the best performance available in April 2024 at cost which is reduced by a factor of 150 (Anthropic Claude 3 Opus, 76\% at \$15/Mtoken). That top performance from April 2024 can now almost be achieved by the open-source Qwen 3 8B model for 3.5¢, representing an over 400x reduction in cost. It's also noteworthy that 8B parameters allows for local use on relatively accessible hardware.  

\subsubsection{Accuracy vs Size: Scaling of Pedagogical Knowledge}

\begin{figure}[htpb]
  \centering
  \includegraphics[width=\linewidth]{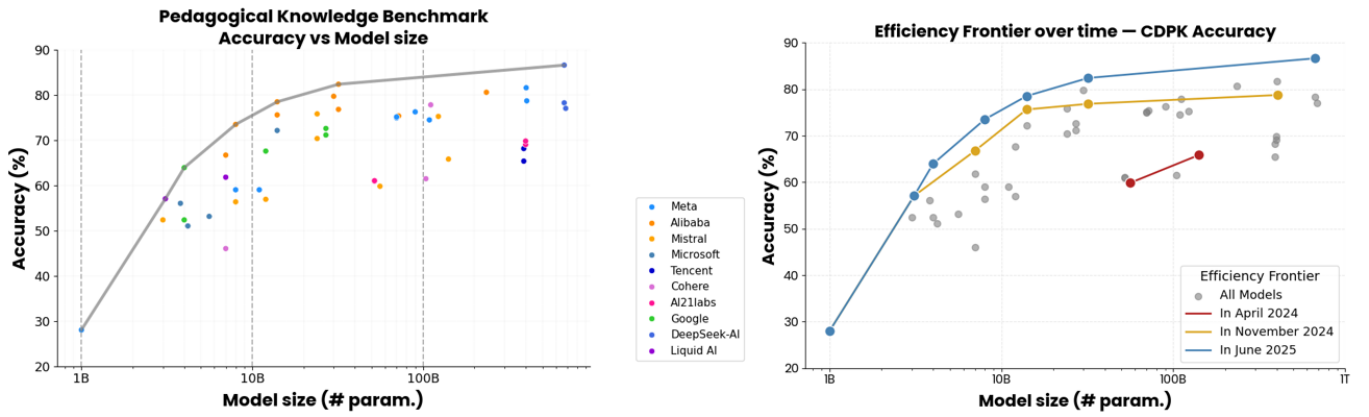}
  \caption{Left: Trade-off between performance and model size, the best models being on the \textit{efficiency frontier} (grey line). Right: Efficiency frontier over time.}
  \label{fig:scatter_cdpk_size}
\end{figure}

Figure \ref{fig:scatter_cdpk_size} shows how performance on our pedagogy benchmarks scales with model size. Note this analysis only includes models for which accurate parameter counts are openly available (i.e. open-weight models). Most of the major closed model providers do not provide architectural details like full parameter counts for their flagship models. As we saw above for cost, the largest model is not the best performing although the smallest model is the worst performing. The Pareto efficiency frontier shows a sharp drop-off below around 8B parameters. However, even at 7B parameters there is a range of performance of over 20 percentage points (from 46\% to 66\%). 

Model size is the crucial parameter which informs the hardware necessary to use the model. Models with more than 70B parameters require multiple high-end GPUs or specialized clusters while 13-70B parameters models require a single high-end GPU (with 40-80GB VRAM, costing tens of thousands of dollars). Models with 7-13B parameters can run on a high-end consumer GPU with 16-24GB VRAM (costing thousands of dollars) while 1-7B parameter models are accessible with gaming computers or laptops with 8-16GB VRAM. Models in the 1-3B parameter range begin to be usable offline on mobile devices. 

The Pareto frontier for size on CDPK includes the models Llama 3.2 1B (Meta), LFM-3B (Liquid AI), Gemma 3n E4B (Google), Qwen 3 8B, 14B and 32B (Alibaba) and Deepseek R1. Gemma 3n E4B is a model optimized for use in everyday devices, such as phones, laptops, and tablets, and again Liquid AI appears with the non-transformer architecture.   

\subsubsection{Variation between Subject Categories}

\begin{figure}[htpb]
  \centering
  \includegraphics[width=\linewidth]{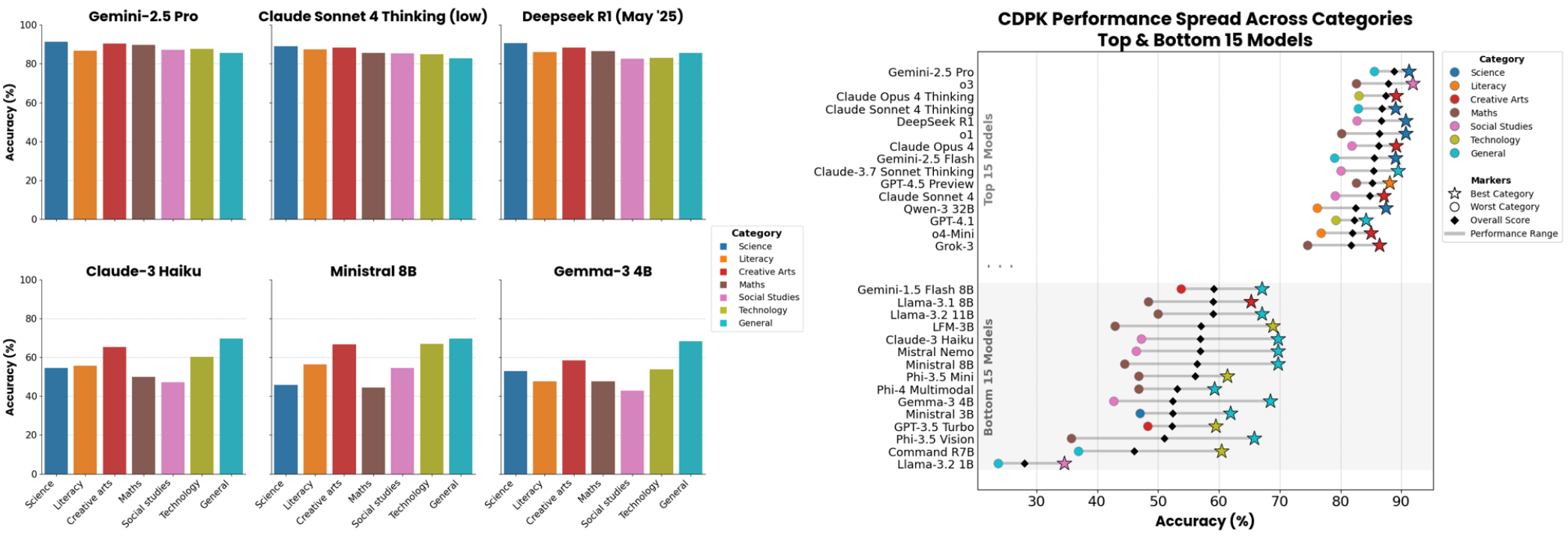}
  \caption{Pedagogy Benchmark (CDPK) accuracy by subject area. Left: accuracy in individual subject areas are shown for three of the 15 top performing models (top) and three of the 15 lowest performing models (bottom). Right: the spread of performance over subjects for the top 15 and bottom 15 performing models. The highest and lowest subject accuracy are indicated by marker colour.}
  \label{fig:cdpk_categories_spread}
\end{figure}

Figure \ref{fig:cdpk_categories_spread} (left panel) shows subject-specific sub-scores for a small selection of the best (top row) and worst (bottom row) performing models. The right panel shows the range of sub-scores over the different subject categories, and shows the top and bottom performing categories for each model. This reveals two key distinctions between model tiers. Firstly, the bottom 15 models demonstrate a much higher performance variance across subjects, evidenced by their wider 'Performance Range' bars. Secondly, these lower-performing models consistently achieve their best scores (indicated by stars) in the Technology and General categories. This specialization is absent in the top 15 models, which show less variance and a greater diversity in their peak-performance subjects. This suggests that the smaller models excel primarily in what may be the least challenging categories, while top models are more versatile.

\subsection{Special Educational Needs and Disabilities (SEND) Pedagogy Benchmark}
\label{sec:send}

As described above, the SEND questions were separated because they involve specialist knowledge as well as teaching and assessment strategies not typically expected for mainstream classroom teachers. While some SEND needs are addressed in regular schools, many of the scenarios in these questions refer to cases managed by specialist staff or in special schools. We therefore evaluated the accuracy of models on the 223 SEND MCQs as a separate SEND-specific pedagogy benchmark\footnote{\url{https://rebrand.ly/sendpedagogy}}.

\begin{table}
    \centering
    \small
    \fontsize{8}{9}\selectfont
    \begin{tabularx}{\textwidth}{X}
        \toprule
        \multicolumn{1}{c}{\textbf{Example}} \\
        \midrule
        Read the following situation:  Rodrigo, a preschool student attending a special language school, is characterized by being participative and restless. It is common for him not to finish tasks during classes, as he gets up from his seat, hides the school supplies of his peers, or starts doing some activity different from what the special education teacher proposed, regularly distracting himself.  Considering the described situation, in which of the following areas should support focus to reduce barriers to learning for the student? \\
        \\
        A) Social\\ 
        B) Health\\ 
        C) Cognitive\\ 
        D) Emotional\\ 
        \\
        Correct answer: C \\
        \bottomrule
    \end{tabularx}
    \caption{Example question from the SEND dataset.}
    \label{tab:example_send_mcqs}
\end{table}

Differences in performance between this benchmark and the original Pedagogy Benchmark underscore the additional challenges associated with SEND-specific pedagogy, as well as the limited availability of high-quality resources in this area. While the results of the two benchmarks were overall quite highly correlated ($r=0.94$), it's interesting that higher performing models (above $~60\%$ accuracy) tended to do better on general pedagogy than SEND, while for lower performing models, this was reversed. The results of all the models can be found in the Appendix (Table \ref{send_accuracy_full}).

\begin{figure}[htbp]
  \centering
  \includesvg[width=\textwidth]{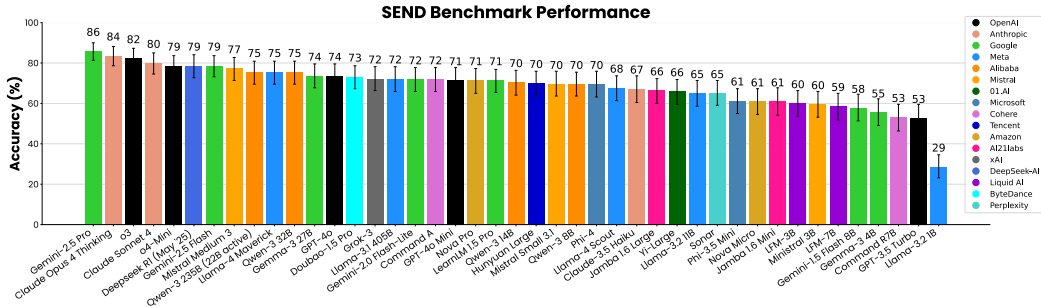}
  \caption{Accuracy on the SEND Benchmark for a subset of models. Error bars show bootstrap 95\% confidence intervals.}
  \label{fig:send_results_overall}
\end{figure}

\begin{figure}[htbp]
  \centering
  \includesvg[scale = 0.3]{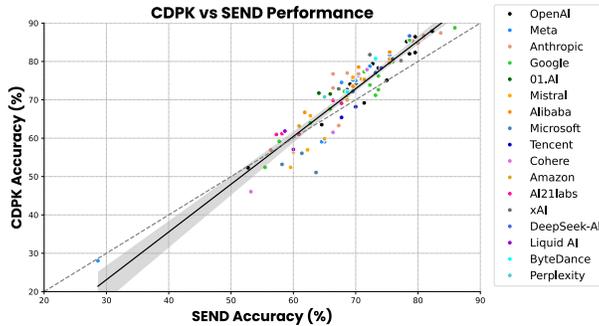}
  \caption{Pedagogy (CDPK) vs SEND Benchmark Performance}
  \label{fig:cdpk_vs_send_scatter}
\end{figure}



%
%
%

%
%


\section{Discussion}

We introduce two novel multiple-choice LLM benchmarks of pedagogical knowledge, based on professional teacher exam questions developed by the \emph{Education Quality Agency and Center for Pedagogical Improvement, Experimentation, and Research} of the Chilean Ministry of Education\footnote{ \url{https://www.cpeip.cl/cpeip/}}. The first, CDPK, focuses on a broad range of general pedagogical knowledge across different educational levels, subjects and pedagogical domains. The second, SEND, focuses on pedagogy related to special educational needs and disabilities. Using genuine teacher exam questions was important to ensure that the benchmark reflects real-world needs and provides an authentic measure of the pedagogical knowledge valued in teacher practice. 

Previous benchmarks focus on content knowledge, measuring what models know about subject matter. Evaluating pedagogical knowledge reveals what models understand about how that content can be effectively taught and assessed. Both of these knowledge components are crucial for the development and selection of effective educational AI tools. For example, it is helpful to understand whether a tool can accurately answer teachers’ queries about science principles used in their lesson. However, determining whether the tool can also provide high-quality, age-appropriate guidance on how these principles can be taught effectively is even more advantageous. The Pedagogy Benchmark results help identify which LLMs are best suited for use in AI educational tools where pedagogical knowledge guides their outputs, including lesson planning assistants, assessment support tools or professional learning platforms.  

We apply these novel benchmarks to a broad range of 97 LLMs, with inference costs ranging from \$0.01-\$100 per million tokens, and sizes ranging from 1B to 700B parameters. Accuracy in CDPK ranges from 28\% (Meta Llama 3.2 1B), to 89\% (Google Gemini 2.5 Pro). For SEND, accuracy ranges from 29\% (Meta Llama 3.2 1B),  to 86\% (Google Gemini 2.5 Pro).  
The best performing models were those specifically fine-tuned for and equipped with \emph{thinking} or \emph{reasoning} functionality. It's interesting that although these were explicitly designed as knowledge benchmarks, reasoning appears to improve performance. It is also noteworthy that the majority of the top performers are proprietary closed-weight models, with Deepseek R1 (an open-weight reasoning model) being the notable exception. 

We also provide inference costs, which allows us to chart the Pareto value frontier. This analysis reveals the models which achieve the best pedagogical knowledge performance at any specific price point. The last 18 months have seen large increases in performance across the full range of inference costs (spanning four orders of magnitude). For example, at \$0.10 per million tokens, the performance has increased from ~50\% (April 2024) to ~70\% (November 2024) to 82\% (June 2025). This analysis also demonstrates that there is no single \emph{best} model; rather, the optimal choice is highly dependent on the specific application's constraints regarding cost, performance requirements, and deployment environment.

The scaling of pedagogical knowledge performance with both inference cost and model size are particularly important when considering applications running on-device and which must support offline use cases. This is important for many LMIC contexts in which internet connectivity may be limited. A notable example of this is the Liquid AI LFM-7B model\footnote{\url{https://www.liquid.ai/blog/introducing-lfm-7b-setting-new-standards-for-efficient-language-models}}, which, with its novel non-transformer architecture, scores 62\% accuracy on CDPK at a very low inference cost of \$0.01 per million tokens. While this is a proprietary model which may not be available for offline on-device deployment, its performance is promising. Similarly, Google's Gemma-3n E4B, an open-weight model explicitly developed for use on local devices such as phones, tablets and laptops scores 64\%. These examples highlight rapid progress and if these trends continue it seems likely that we will have on-device models which can saturate the pedagogy benchmarks within a relatively short timeframe. 

As question-level human performance data is not available, we estimate approximate human accuracy based on the past performance of Chilean teachers on complete exams to be approximately 50\%. It is crucial to recognise, however, that these benchmarks exclusively assess pedagogical knowledge. These benchmarks do not, and cannot, measure the many other critical practical dimensions of teaching, such as classroom management, student-teacher relationships, and the ability to inspire and motivate learners, all of which are fundamental to effective teaching practice. 

\subsection{Limitations}

We have provided bootstrap confidence intervals around the accuracy scores, and investigated reliability over multiple repeated runs. This shows care must be taken interpreting the precise rank ordering, as the difference between the first place and second place model is less than 1\% which may not be meaningful. However, over the full range of performance there are clearly robust differences between models. 

We have relied on a single source of questions for these benchmarks - the Chilean Ministry of Education. These are a comprehensive, broad,  well-researched and designed set of exams. Care was taken to remove any questions judged to be too specific to the Chilean context or requiring specific knowledge about Chile. Nevertheless, pedagogy is not universal, and the principles, priorities and practices valued in the Chilean educational system may differ from those in other countries or cultural contexts. Additionally, all our questions have been translated from Spanish into English. Translation can sometimes lose subtle nuances. Key pedagogical terms or situational descriptions might carry slightly different connotations in English, which could affect models' interpretation of the question. We did search for other similar exams, but many countries do not have formal exams covering pedagogical knowledge, opting instead for experiential training and live classroom assessment. We did find some similar exams from other countries, but these were not available under an open-license so could not be used in a public dataset of the sort we aimed to produce here.

Our approach uses a single standardised prompt to ensure fair comparison between all models. But this may not best capture the capabilities of all models. For example, we observe thinking models, which have a separate output space for reasoning, generally perform better. It's possible an alternative chain-of-thought prompting approach that invited non-reasoning models to think carefully about the options step-by-step rather than answering directly could improve performance. Similarly, we use a few-shot prompt with three example questions. While this is recommended for non-reasoning models, it may not be optimal for reasoning models. The particular few-shot examples were fixed (for each subject), for replicability and consistency, but different selections of examples might differentially affect performance of different models. 

As our dataset has not been publicly released prior to this paper, we believe that the results presented here should be free of contamination or leakage. While some of the source exam papers are available online, they were in Spanish without correct answers directly linked. Contamination may be an issue for the future, but following standard practice we request users of our dataset not to reveal examples from the dataset online. 

The SEND and CDPK benchmark results are highly correlated. We have not explored here the degree to which their results might also be directly correlated to other benchmarks like MMLU or GPQA. We therefore cannot be sure of the degree to which we are isolating specific knowledge of pedagogy, versus evaluating a model's overall knowledge base and ability to answer factual MCQs. However, we believe our new benchmarks are useful for several reasons. We provide online leaderboards which we update with new models, and we present inference costs alongside the models to allow direct value judgements. Even if they are correlated to other knowledge benchmarks our questions are directly testing pedagogy which is important for educational applications. Further, as above, these benchmarks are likely not contaminated by having models fine-tuned directly on the questions. We hope they will be particularly useful for informing the choice of smaller and cheaper models for educational applications in resource-limited contexts. 

\subsection{Future Directions}

We highlight the need to move beyond general knowledge benchmarks and focus on specialized domains like education. We see our pedagogical knowledge benchmarks as a small step in the path towards a robust ecosystem of education-related evaluations of generative AI models and products. For model developers, developments in this area can guide fine-tuning efforts to improve pedagogical practice and enhance outcomes for learners globally. For EdTech product developers, our leaderboards and future work in this direction can serve as a practical guide for model selection, enabling them to make informed decisions that balance pedagogical performance with the inference cost and latency constraints of their specific tools.

While MCQs are a well-established approach for standardized knowledge assessment, they do not evaluate an LLM's generative or interactive capabilities. For instance, a benchmark question can assess if a model can identify the best teaching strategy from a list, but not how well it can generate a novel lesson plan, create differentiated materials, or scaffold a concept in a conversational turn-by-turn interaction with a student or teacher. As noted in the Introduction, this distinction between pedagogical knowledge and pedagogical practice is crucial to keep in mind. Here we explicitly focussed on pedagogical knowledge as it can be more easily tested across a wide range of language models. Considering how this can be contextualised and extended to more realistic evaluations of models' real-world pedagogical practice is an important area for future work. This could involve designing new evaluation paradigms that move beyond static, multiple-choice formats. Future benchmarks could involve scenario-based assessments where models must generate open-ended responses to complex classroom situations, with evaluation conducted via expert-designed rubrics. Further, developing simulated classroom interactions could test an AI's ability to respond dynamically and appropriately to student or teacher inputs.

While acknowledging the difference between knowledge and practice, it is also possible that these are related in LLMs. For teachers, there is evidence that pedagogical knowledge aligns with many practical measures of teaching quality \cite{reynolds_examining_2021}. Similarly, a model that produces pedagogically sound behaviour is likely to be able to answer pedagogical knowledge questions about that behaviour. Although we do not test it explicitly, we hope our benchmarks provide useful information that is related to pedagogical practice in real applications. We hope that future work will compare our benchmarks to direct measures of models' behaviour to elucidate any such relationship.

Finally, as models approach, and in many cases exceed, human performance on these knowledge benchmarks, a dedicated research agenda into their responsible and ethical deployment is essential. The increasing capability of these models brings risks of misuse, such as promoting over-reliance among novice teachers, creating an illusion of expertise that could lead to harm if incorrect advice is followed, or propagating the subtle cultural and pedagogical biases inherent in their training data. Future work must therefore investigate the development of clear ethical guardrails, robust user training protocols, and "human-in-the-loop" systems to ensure that these powerful tools are deployed safely and effectively, augmenting teacher professionalism rather than attempting to replace it.

\subsection{Conclusion}

This paper establishes a foundational benchmark for LLM pedagogical knowledge, revealing that models are rapidly mastering the principles of effective teaching. Yet, this development highlights an important challenge: the distinction between a theoretical knowledge of teaching and its translation into effective classroom practice. Therefore, the central task for the field is twofold: to continue building these vital, education-focused benchmarks including those which assess practical application of knowledge, while simultaneously working to gather supporting evidence of impact from real world testing. Only by addressing both can we ensure AI serves as a truly effective and responsible tool for teachers and learners across the globe.


\section*{Acknowledgements}

We thank the Agencia de la Calidad de Educación y Centro de Perfeccionamiento, Experimentación e Investigaciones Pedagógicas (Education Quality Agency and Center for Pedagogical Improvement, Experimentation, and Research), \url{https://www.cpeip.cl/cpeip/} of the Chilean Ministry of Education for developing and sharing the exams we sourced questions from. 

Funding for the AI-for-Education initiative has been provided by Jacobs Foundation, the Foreign Commonwealth and Development Office (FCDO) of the UK Government, and the Gates Foundation. Any opinions, findings, and conclusions or recommendations expressed in this material are those of the authors and do not necessarily reflect the views of these funders.


\bibliography{citations}
\newpage


\appendix

\section{Example Questions}

\begin{table}[H]
    \centering
    \fontsize{8}{9}\selectfont
    \begin{tabularx}{\textwidth}{X}
        \toprule
        \multicolumn{1}{c}{\textbf{Maths (Pre-primary)}} \\
        \midrule
        In the core of logical-mathematical relationships and quantification, a preschool educator conducts the following learning experience with the children: in small groups, she asks them to take two bars of plasticine with the same amount, to make a big ball with one and small balls with the other. Then, she asks the children: "Where do you think there is more plasticine?". Considering the presented situation, which logical-mathematical reasoning skill is promoted with this activity? \\
        \\
        A) Estimation\\ 
        B) Conservation\\ 
        C) Decomposition\\ 
        D) Correspondence\\ 
        \\
        Correct answer: B \\
        \midrule
        \multicolumn{1}{c}{\textbf{Literacy (Primary)}} \\
        \midrule
        A second-grade teacher worked on the story of Cinderella with their class and then conducted a brief assessment of the students' levels of understanding. One of the questions the teacher asked was: “Why did Cinderella's stepsisters insist on trying on the glass slipper even though it didn’t fit them well?”  What evaluation indicator is addressed through the question asked by the teacher? \\
        \\
        A) They recognize the conflict in narrative texts \\
        B) They establish cause relationships in narrative texts \\
        C) They identify sequences of events in narrative texts \\
        D) They identify characteristics of secondary characters in narrative texts \\
        \\
        Correct answer: B \\
        \midrule
        \multicolumn{1}{c}{\textbf{Creative arts (Secondary)}} \\
        \midrule
        Which of the following evaluation activities, conducted before building the group installation, would help students recognize the level of achievement of their projects? \\
        \\
        A) They appreciate, through a dialogue, the sketches of their projects, discussing the proposed visual language \\ 
        B) They analyze their projects, identifying the coherence between their ideas, materiality, and the selected visual language \\ 
        C) They identify the materials used in contemporary artists' installations and contrast them with those selected in their projects \\ 
        D) They learn about installation projects carried out in other establishments and analyze them from the visual language perspective \\ 
        \\
        Correct answer: B \\
        \midrule
        \multicolumn{1}{c}{\textbf{Technology (Secondary)}} \\
        \midrule
        In a first-year high school class, the Technology teacher asks the students to develop a project called "Emprendemás." In this project, the students must create a company that offers a service to address a need of a specific segment of society. When the students present their ideas explaining the need they will address and the service they will offer, the teacher notices that some teams are unable to define the market segment to which the service is directed. 
        What activity could the teacher propose so that the students can determine the segment for which the service they want to offer is destined? \\
        \\
        A) Ask the students to analyze the characteristics of the people in whom they detected the need so that they can systematize the information, and then create an empathy map to understand what they want \\
        B) Ask the teams to watch a video showing the services offered by the competition and determine the type of clients they serve, and then compare them with the potential recipients of their product \\
        C) Ask the teams to investigate which social segments the competition offers services like those they propose, so that they can analyze them and determine how they differ, defining the value proposition of the service they are offering \\
        D) Ask the students to create and send an online survey to different people asking if they would use the service they are proposing, and collect data from the respondents, such as age, salary range, profession, nationality, and gender \\
        \\
        Correct answer: A \\
        \bottomrule
    \end{tabularx}
    \caption{Example questions from the dataset.}
    \label{tab:display_questions}
\end{table}

\section{Data Extraction Prompts}

\begin{tcolorbox}[title=Data Extraction \& Translation Prompt]
    \small
    \textbf{You are a highly skilled translator and text processor. I have a text containing paragraphs of educational questions and answers in Spanish. I want you to do the following:\\
    \\
    1. **Segment** each paragraph into "Question" and "Answers":\\
    - "Question" should contain the main body of the paragraph that describes the situation and the instruction. Basically, the part before the list of choices. If there is a paragraph of text before the actual question, include that as well.\\
    - "Answers" should include the list of choices (e.g., A, B, C, D) provided after the question.\\
    \\
    2. **Translate** the segmented text into English.\\
    \\
    The text is as follows:}
    \\
    -----\\
    {paragraph}\\
    -----\\
    \\
    \textbf{1. Please provide the segmented output in the following JSON format:\\
    {{\\
        "question": "<Segmented Question>",\\
        "A": "<Segmented Choice A>",\\
        "B": "<Segmented Choice B>",\\
        "C": "<Segmented Choice C>",\\
        "D": "<Segmented Choice D>"\\
    }}\\
    \\
    2. Please provide the translated output in the following JSON format:\\
    {{\\
        "question": "<Translated Question>",\\
        "A": "<Translated Choice A>",\\
        "B": "<Translated Choice B>",\\
        "C": "<Translated Choice C>",\\
        "D": "<Translated Choice D>"\\}
    }}\\
    \label{prompt_data_extraction}
\end{tcolorbox}

\begin{tcolorbox}[title=Prompt OCR]
    \small
    \textbf{You are a specialized OCR system trained to deliver highly accurate text extraction. Please extract all visible 
    text in each image with precision, including details like formatting, punctuation, and capitalization. 
    Maintain the text's exact sequence as shown in the image without adding or omitting any content. 
    Provide only the text as output—no additional comments, descriptions, or formatting adjustments.}
    \label{prompt_ocr}
\end{tcolorbox}

\section{Benchmark Question Prompt}
\begin{tcolorbox}[title=Prompt]
    \small
    The following are example multiple choice questions (with answers).\\
    
    A preschool educator is working on the following learning objective from the Verbal Language core: "Understand, through attentive listening, explicit content from literary and non-literary texts, recognizing central ideas, indicating preferences, making simple descriptions, asking about the content." To conduct a process evaluation, the following achievement indicator is set: Understands explicit content from literary texts.  Which of the following activities is appropriate to obtain evidence of the level of achievement of the indicated indicator?\\
    A. The educator shows the children the cover of the fable "The Hare and the Tortoise" and then asks them what they think this text will be about\\
    B. The educator invites the children to act out the fable "The Lion and the Mouse" and then asks them about the lesson the story and its characters teach\\
    C. The educator tells the story "The Color Monster" and then asks the children what colors the emotions had that the Color Monster could recognize\\
    D. The educator invites the children to dramatize the story "No Laughing, Pepe" and then asks them what part they liked most about the story and encourages them to create another ending to the story\\
    Correct Answer: C\\
    \\
    A second-grade teacher worked on the story of Cinderella with their class and then conducted a brief assessment of the students' levels of understanding. One of the questions the teacher asked was: “Why did Cinderella's stepsisters insist on trying on the glass slipper even though it didn’t fit them well?”  What evaluation indicator is addressed through the question asked by the teacher?\\
    A. They recognize the conflict in narrative texts \\
    B. They establish cause relationships in narrative texts \\
    C. They identify sequences of events in narrative texts \\
    D. They identify characteristics of secondary characters in narrative texts \\
    Correct Answer: B \\
    \\
    A teacher is working on the inferential level of reading comprehension based on the reading of the following short story: Vacations. Next year my family plans to go to the moon. My mother told me to invite a friend if I wanted, of course, to not get bored she must think. The truth is that I'm not very excited, I don't know, I've never liked going off the planet much, I prefer to eat noodles with sauce and cheese. Anyway, my friend says she would love to go, but tells me she needs a new bikini, they say women on the moon are very beautiful, but I don't believe it. We will stay at the Hotel Armstrong and travel on Pullmanmoon (hopefully it has a bathroom). I'll bring a melon. (Catalina Yáñez) What question could the teacher ask the students to promote reading at the inferential level? \\
    A. What does the protagonist feel about the trip to the moon? \\
    B. What should a vacation trip to the moon be like? \\
    C. Why does the protagonist's mother tell her to invite a friend? \\
    D. What attitude does the protagonist take regarding the trip at the end of the story? \\
    Correct Answer: D\\
    \\
    Answer the following real question using same answer format:
    \\
    \\
    This model is based on learning to read and write progressively, step by step, with explicit teaching of letters, types of syllables, words, sentences, and finally texts, starting from mastering the code to understanding more complex texts, that is, it is framed within the processes of decoding reading. Which reading teaching model is this methodological approach related to? \\
    A. Integrated \\
    B. Holistic \\
    C. Skills-based \\
    D. Transactional \\
    \\
    Only answer the real question.\\
    Only provide the letter for your answer. \\
    Stop exactly after the letter.
    \label{prompt_example}
\end{tcolorbox}

\section{Full results}
\label{sec:full_results}

\begin{table}[H]
    \centering
    \resizebox{\textwidth}{!}{%
    \begin{tabular}{l c l| l c l}
    \toprule
    Model & Accuracy & Company & Model & Accuracy & Company \\
    \midrule
    Gemini-2.5 Pro & 88.77 & \textcolor{gray}{Google} & Gemini-1.5 Pro & 73.86 & \textcolor{gray}{Google} \\
    o3 & 87.88 & \textcolor{gray}{OpenAI} & Qwen-3 8B & 73.53 & \textcolor{gray}{Alibaba} \\
    Claude Opus 4 Thinking (low) & 87.43 & \textcolor{gray}{Anthropic} & Claude-3.5 Sonnet June & 73.08 & \textcolor{gray}{Anthropic} \\
    Claude Sonnet 4 Thinking (low) & 86.76 & \textcolor{gray}{Anthropic} & Grok Beta & 72.86 & \textcolor{gray}{xAI} \\
    Deepseek R1 (May '25) & 86.65 & \textcolor{gray}{DeepSeek-AI} & Gemma-3 27B & 72.64 & \textcolor{gray}{Google} \\
    o1 & 86.43 & \textcolor{gray}{OpenAI} & o1-Mini & 72.64 & \textcolor{gray}{OpenAI} \\
    Claude Opus 4 & 86.32 & \textcolor{gray}{Anthropic} & Phi-4 & 72.19 & \textcolor{gray}{Microsoft} \\
    Gemini-2.5 Flash & 85.54 & \textcolor{gray}{Google} & Gemini-1.5 Flash & 72.19 & \textcolor{gray}{Google} \\
    Claude-3.7 Sonnet Thinking (medium) & 85.43 & \textcolor{gray}{Anthropic} & Doubao-1.5 Lite & 72.08 & \textcolor{gray}{ByteDance} \\
    GPT-4.5 Preview & 85.21 & \textcolor{gray}{OpenAI} & Yi Lightning & 71.75 & \textcolor{gray}{01.AI} \\
    Claude Sonnet 4 & 84.76 & \textcolor{gray}{Anthropic} & Yi-Large & 71.52 & \textcolor{gray}{01.AI} \\
    Qwen-3 32B & 82.42 & \textcolor{gray}{Alibaba} & Gemma-2 27B & 71.19 & \textcolor{gray}{Google} \\
    GPT-4.1 & 82.31 & \textcolor{gray}{OpenAI} & Sonar & 70.75 & \textcolor{gray}{Perplexity} \\
    o4-Mini & 81.98 & \textcolor{gray}{OpenAI} & Mistral Small 3 & 70.41 & \textcolor{gray}{Mistral} \\
    Grok-3 & 81.76 & \textcolor{gray}{xAI} & Nova Lite & 69.97 & \textcolor{gray}{Amazon} \\
    Llama-4 Maverick & 81.65 & \textcolor{gray}{Meta} & Jamba 1.6 Large & 69.86 & \textcolor{gray}{AI21labs} \\
    Doubao-1.5 Pro & 80.76 & \textcolor{gray}{ByteDance} & GPT-4o Mini & 69.19 & \textcolor{gray}{OpenAI} \\
    Grok-3 Mini & 80.65 & \textcolor{gray}{xAI} & Jamba 1.5 Large & 69.08 & \textcolor{gray}{AI21labs} \\
    Qwen-3 235B (22B active) & 80.65 & \textcolor{gray}{Alibaba} & Hunyuan Large & 68.19 & \textcolor{gray}{Tencent} \\
    Grok-2 & 80.20 & \textcolor{gray}{xAI} & Gemma-3 12B & 67.63 & \textcolor{gray}{Google} \\
    Mistral Medium 3 & 79.98 & \textcolor{gray}{Mistral} & Qwen-2.5 7B & 66.74 & \textcolor{gray}{Alibaba} \\
    Gemini-2.0 Flash & 79.87 & \textcolor{gray}{Google} & Mixtral-8x22B & 65.85 & \textcolor{gray}{Mistral} \\
    Qwen-3 30B (3B active) & 79.76 & \textcolor{gray}{Alibaba} & Hunyuan Large (Long Context) & 65.41 & \textcolor{gray}{Tencent} \\
    o3-Mini & 79.42 & \textcolor{gray}{OpenAI} & Gemma-3n E4B & 63.96 & \textcolor{gray}{Google} \\
    Llama-3.1 405B & 78.75 & \textcolor{gray}{Meta} & GPT-4.1 Nano & 63.52 & \textcolor{gray}{OpenAI} \\
    Qwen-3 14B & 78.53 & \textcolor{gray}{Alibaba} & Claude-3.5 Haiku & 63.29 & \textcolor{gray}{Anthropic} \\
    Deepseek V3 & 78.31 & \textcolor{gray}{DeepSeek-AI} & Nova Micro & 63.18 & \textcolor{gray}{Amazon} \\
    GPT-4o & 78.31 & \textcolor{gray}{OpenAI} & LFM-7B & 61.85 & \textcolor{gray}{Liquid AI} \\
    Command A & 77.86 & \textcolor{gray}{Cohere} & Command-R+ August & 61.51 & \textcolor{gray}{Cohere} \\
    Gemini-2.0 Flash-Lite & 77.64 & \textcolor{gray}{Google} & Jamba Instruct & 61.18 & \textcolor{gray}{AI21labs} \\
    LearnLM 1.5 Pro & 77.31 & \textcolor{gray}{Google} & Jamba 1.6 Mini & 61.07 & \textcolor{gray}{AI21labs} \\
    Deepseek V3 0324 & 77.09 & \textcolor{gray}{DeepSeek-AI} & Jamba 1.5 Mini & 60.96 & \textcolor{gray}{AI21labs} \\
    Claude-3 Opus & 76.97 & \textcolor{gray}{Anthropic} & Mixtral-8x7B & 59.84 & \textcolor{gray}{Mistral} \\
    Qwen-2.5 32B & 76.86 & \textcolor{gray}{Alibaba} & Gemini-1.5 Flash 8B & 59.18 & \textcolor{gray}{Google} \\
    Claude-3.5 Sonnet October & 76.75 & \textcolor{gray}{Anthropic} & Llama-3.1 8B & 59.07 & \textcolor{gray}{Meta} \\
    Claude-3.7 Sonnet & 76.75 & \textcolor{gray}{Anthropic} & Llama-3.2 11B & 59.07 & \textcolor{gray}{Meta} \\
    Llama-3.2 90B & 76.31 & \textcolor{gray}{Meta} & LFM-3B & 57.06 & \textcolor{gray}{Liquid AI} \\
    Gemini-2.5 Flash-Lite Preview & 76.20 & \textcolor{gray}{Google} & Claude-3 Haiku & 56.95 & \textcolor{gray}{Anthropic} \\
    Mistral Small 3.1 24B & 75.86 & \textcolor{gray}{Mistral} & Mistral Nemo & 56.95 & \textcolor{gray}{Mistral} \\
    Qwen-2.5 14B & 75.64 & \textcolor{gray}{Alibaba} & Ministral 8B & 56.40 & \textcolor{gray}{Mistral} \\
    Qwen-2.5 72B & 75.42 & \textcolor{gray}{Alibaba} & Phi-3.5 Mini & 56.06 & \textcolor{gray}{Microsoft} \\
    Nova Pro & 75.31 & \textcolor{gray}{Amazon} & Phi-4 Multimodal & 53.17 & \textcolor{gray}{Microsoft} \\
    Mistral Large November & 75.31 & \textcolor{gray}{Mistral} & Ministral 3B & 52.39 & \textcolor{gray}{Mistral} \\
    Llama-3.3 70B & 75.19 & \textcolor{gray}{Meta} & Gemma-3 4B & 52.39 & \textcolor{gray}{Google} \\
    GPT-4.1 Mini & 75.08 & \textcolor{gray}{OpenAI} & GPT-3.5 Turbo & 52.28 & \textcolor{gray}{OpenAI} \\
    Llama-3.1 70B & 74.97 & \textcolor{gray}{Meta} & Phi-3.5 Vision & 51.06 & \textcolor{gray}{Microsoft} \\
    Llama-4 Scout & 74.53 & \textcolor{gray}{Meta} & Command R7B & 46.05 & \textcolor{gray}{Cohere} \\
    GPT-4 Turbo & 74.30 & \textcolor{gray}{OpenAI} & Llama-3.2 1B & 28.03 & \textcolor{gray}{Meta} \\
    GPT-4 & 74.08 & \textcolor{gray}{OpenAI} &  &  &  \\
    \bottomrule
    \end{tabular}
    }
    \caption{Accuracy scores on The Pedagogy Benchmark, sorted from highest (top left) to lowest (bottom right).}
    \label{cdpk_accuracy_full}
\end{table}

\begin{table}[H]
    \centering
    \resizebox{\textwidth}{!}{%
    \begin{tabular}{l c l| l c l}
    \toprule
    Model & Accuracy & Company & Model & Accuracy & Company \\
    \midrule
    Gemini-2.5 Pro & 85.91 & \textcolor{gray}{Google} & Llama-3.3 70B & 69.55 & \textcolor{gray}{Meta} \\
    Claude Opus 4 Thinking (low) & 83.64 & \textcolor{gray}{Anthropic} & Mistral Small 3.1 24B & 69.55 & \textcolor{gray}{Mistral} \\
    o3 & 82.27 & \textcolor{gray}{OpenAI} & GPT-4 & 69.09 & \textcolor{gray}{OpenAI} \\
    Claude Sonnet 4 Thinking (low) & 80.91 & \textcolor{gray}{Anthropic} & Claude-3 Opus & 68.64 & \textcolor{gray}{Anthropic} \\
    Claude Opus 4 & 80.91 & \textcolor{gray}{Anthropic} & Mistral Small 3 & 68.64 & \textcolor{gray}{Mistral} \\
    Claude Sonnet 4 & 80.00 & \textcolor{gray}{Anthropic} & Nova Lite & 68.64 & \textcolor{gray}{Amazon} \\
    o1-Medium & 79.55 & \textcolor{gray}{OpenAI} & o1-Mini & 68.64 & \textcolor{gray}{OpenAI} \\
    GPT-4.1 & 79.55 & \textcolor{gray}{OpenAI} & Doubao-1.5 Lite & 68.64 & \textcolor{gray}{ByteDance} \\
    Claude-3.7 Sonnet Thinking (medium) & 79.09 & \textcolor{gray}{Anthropic} & Gemini-1.5 Flash & 68.18 & \textcolor{gray}{Google} \\
    Gemini-2.5 Flash & 78.64 & \textcolor{gray}{Google} & Llama-4 Scout & 67.73 & \textcolor{gray}{Meta} \\
    Deepseek R1 (May '25) & 78.64 & \textcolor{gray}{DeepSeek-AI} & Jamba 1.5 Large & 67.73 & \textcolor{gray}{AI21labs} \\
    o4-Mini & 78.64 & \textcolor{gray}{OpenAI} & Hunyuan Large (Long Context) & 67.73 & \textcolor{gray}{Tencent} \\
    GPT-4.5 Preview & 78.18 & \textcolor{gray}{OpenAI} & Claude-3.5 Haiku & 67.27 & \textcolor{gray}{Anthropic} \\
    Grok-2 & 77.27 & \textcolor{gray}{xAI} & Grok Beta & 67.27 & \textcolor{gray}{xAI} \\
    Mistral Medium 3 & 77.27 & \textcolor{gray}{Mistral} & Command-R+ August & 66.36 & \textcolor{gray}{Cohere} \\
    Grok-3 Mini & 75.91 & \textcolor{gray}{xAI} & Jamba 1.6 Large & 66.36 & \textcolor{gray}{AI21labs} \\
    Gemini-2.0 Flash & 75.91 & \textcolor{gray}{Google} & Claude-3.5 Sonnet October & 66.36 & \textcolor{gray}{Anthropic} \\
    Llama-4 Maverick & 75.45 & \textcolor{gray}{Meta} & Claude-3.5 Sonnet June & 66.36 & \textcolor{gray}{Anthropic} \\
    Qwen-3 32B & 75.45 & \textcolor{gray}{Alibaba} & Gemma-3 12B & 65.91 & \textcolor{gray}{Google} \\
    Qwen-3 30B (3B active) & 75.45 & \textcolor{gray}{Alibaba} & Yi-Large & 65.91 & \textcolor{gray}{01.AI} \\
    Qwen-3 235B (22B active) & 75.45 & \textcolor{gray}{Alibaba} & Mixtral-8x7B & 65.00 & \textcolor{gray}{Mistral} \\
    GPT-4.1 Mini & 75.00 & \textcolor{gray}{OpenAI} & Llama-3.2 11B & 65.00 & \textcolor{gray}{Meta} \\
    Deepseek V3 & 74.09 & \textcolor{gray}{DeepSeek-AI} & Sonar & 65.00 & \textcolor{gray}{Perplexity} \\
    Gemini-2.5 Flash-Lite Preview & 73.64 & \textcolor{gray}{Google} & Llama-3.1 8B & 64.55 & \textcolor{gray}{Meta} \\
    Gemma-3 27B & 73.64 & \textcolor{gray}{Google} & GPT-4.1 Nano & 64.55 & \textcolor{gray}{OpenAI} \\
    GPT-4o & 73.64 & \textcolor{gray}{OpenAI} & Yi Lightning & 64.09 & \textcolor{gray}{01.AI} \\
    Deepseek V3 0324 & 73.18 & \textcolor{gray}{DeepSeek-AI} & Phi-3.5 Vision & 63.64 & \textcolor{gray}{Microsoft} \\
    Gemma-2 27B & 73.18 & \textcolor{gray}{Google} & Mixtral-8x22B & 62.73 & \textcolor{gray}{Mistral} \\
    Doubao-1.5 Pro & 73.18 & \textcolor{gray}{ByteDance} & Gemma-3n E4B & 62.73 & \textcolor{gray}{Google} \\
    o3-Mini & 72.73 & \textcolor{gray}{OpenAI} & Mistral Nemo & 62.27 & \textcolor{gray}{Mistral} \\
    Llama-3.1 405B & 72.27 & \textcolor{gray}{Meta} & Qwen-2.5 7B & 61.82 & \textcolor{gray}{Alibaba} \\
    Grok-3 & 72.27 & \textcolor{gray}{xAI} & Phi-3.5 Mini & 61.36 & \textcolor{gray}{Microsoft} \\
    Gemini-1.5 Pro & 72.27 & \textcolor{gray}{Google} & Jamba 1.6 Mini & 60.91 & \textcolor{gray}{AI21labs} \\
    Command A & 71.82 & \textcolor{gray}{Cohere} & Nova Micro & 60.91 & \textcolor{gray}{Amazon} \\
    Gemini-2.0 Flash-Lite & 71.82 & \textcolor{gray}{Google} & LFM-3B & 60.00 & \textcolor{gray}{Liquid AI} \\
    Mistral Large November & 71.36 & \textcolor{gray}{Mistral} & Ministral 8B & 60.00 & \textcolor{gray}{Mistral} \\
    LearnLM 1.5 Pro & 71.36 & \textcolor{gray}{Google} & Ministral 3B & 59.55 & \textcolor{gray}{Mistral} \\
    Nova Pro & 71.36 & \textcolor{gray}{Amazon} & LFM-7B & 58.64 & \textcolor{gray}{Liquid AI} \\
    GPT-4o Mini & 71.36 & \textcolor{gray}{OpenAI} & Jamba Instruct & 58.18 & \textcolor{gray}{AI21labs} \\
    Qwen-2.5 72B & 70.91 & \textcolor{gray}{Alibaba} & Phi-4 Multimodal & 58.18 & \textcolor{gray}{Microsoft} \\
    Claude-3.7 Sonnet & 70.45 & \textcolor{gray}{Anthropic} & Gemini-1.5 Flash 8B & 57.73 & \textcolor{gray}{Google} \\
    Llama-3.2 90B & 70.45 & \textcolor{gray}{Meta} & Jamba 1.5 Mini & 57.27 & \textcolor{gray}{AI21labs} \\
    Qwen-3 14B & 70.45 & \textcolor{gray}{Alibaba} & Claude-3 Haiku & 56.36 & \textcolor{gray}{Anthropic} \\
    Hunyuan Large & 70.00 & \textcolor{gray}{Tencent} & Gemma-3 4B & 55.45 & \textcolor{gray}{Google} \\
    Llama-3.1 70B & 70.00 & \textcolor{gray}{Meta} & Command R7B & 53.18 & \textcolor{gray}{Cohere} \\
    GPT-4 Turbo & 70.00 & \textcolor{gray}{OpenAI} & GPT-3.5 Turbo & 52.73 & \textcolor{gray}{OpenAI} \\
    Qwen-3 8B & 69.55 & \textcolor{gray}{Alibaba} & Llama-3.2 1B & 28.64 & \textcolor{gray}{Meta} \\
    Phi-4 & 69.55 & \textcolor{gray}{Microsoft} & &  &  \\
    \bottomrule
    \end{tabular}
    }
    \caption{Accuracy scores on The SEND Pedagogy Benchmark, sorted from highest (top left) to lowest (bottom right).}
    \label{send_accuracy_full}
\end{table}

\end{document}